\documentclass[sigconf]{acmart}
\usepackage{subfigure}
\usepackage{multicol}
\AtBeginDocument{%
  }

\setcopyright{acmlicensed}
\copyrightyear{2024}
\acmYear{2024}
\acmDOI{XXXXXXX.XXXXXXX}

\copyrightyear{2024}
\acmYear{2024}
\setcopyright{acmcopyright}\acmConference[HASCA '24]{Human Activity Sensing Corpus and Applications}{October 7-9 2024}{Melbourne, Australia}

\definecolor{ao(english)}{rgb}{0.0, 0.5, 0.0}
\definecolor{awesome}{rgb}{1.0, 0.13, 0.32}

\newcommand{\thomas}[1]{\textbf{\sffamily{\textcolor{brown}{[#1 -- Thomas]}}}}
\definecolor{ao(english)}{rgb}{0.0, 0.5, 0.0}
\newcommand{\positive}[1]{{\textcolor{ao(english)}{#1}}}
\newcommand{\negative}[1]{{\textcolor{red}{#1}}}
\begin{document}

\title{Game of LLMs: Discovering Structural Constructs in Activities using Large Language Models}

\author{Shruthi K. Hiremath}
\email{shiremath9@gatech.edu}
\orcid{0000-0003-4317-8864}
\affiliation{%
  \institution{School of Interactive Computing, Georgia Institute of Technology, Atlanta}
  \streetaddress{E1566B. CODA 15th Floor, Georgia Institute of Technology}
  \city{Atlanta}
  \state{Georgia}
  \country{USA}
  \postcode{30308}
}

\author{Thomas Pl{\"o}tz}
\email{thomas.ploetz@gatech.edu}
\orcid{0000-0002-1243-7563}
\affiliation{%
  \institution{School of Interactive Computing, Georgia Institute of Technology, Atlanta}
  \streetaddress{E1564B. CODA 15th Floor, Georgia Institute of Technology}
  \city{Atlanta}
  \state{Georgia}
  \country{USA}
  \postcode{30308}
}


\begin{abstract}
Human Activity Recognition is a time-series analysis problem. 
A popular analysis procedure used by the community assumes an optimal window length to design recognition pipelines.
However, in the scenario of smart homes, where activities are of varying duration and frequency, the assumption of a constant sized window does not hold. 
Additionally, previous works have shown these activities to be made up of building blocks. 
We focus on identifying these underlying building blocks--structural constructs, with the use of large language models.
Identifying these constructs can be beneficial especially in recognizing short-duration and infrequent activities.
We also propose the development of an  activity recognition procedure that uses these building blocks to model activities, thus helping the downstream task of activity monitoring in smart homes. 
\end{abstract}



\received{14 June 2024}
\received[revised]{14 June 20249}
\received[accepted]{14 June 2024}

\maketitle

\section{Introduction}
\label{sec1-introduction}
Developing human activity recognition systems for ambient settings--such as smart homes--is essential to providing assistance, aid and support to residents \cite{asadzadeh2022review,bouchabou2021survey}. 
Prior works have shown that such applications provide support for diverse populations -- from monitoring activities of the elderly population to technologically assisting the ``sandwich generation" in their daily life \cite{majumder2017smart,pal2018internet,burke2017sandwich,pashazade2023explaining}. 
With the decrease in automation costs and ease of instrumenting smart homes, collecting data in these environments has become a possibility for many \cite{crandall2012smart, gazis2021smart, belley2014efficient}.
Advancements in the design of methodologies to analyze data
provide for robust recognition pipelines \cite{guan2017ensembles,deldari2022beyond,tang2021selfhar}.  

Challenges exist in analyzing data collected in smart homes. 
A popular approach in the wearable community (data with constant sampling rates) is to use the Activity Recognition Chain (ARC) \cite{bulling2014tutorial,plotz2021applying} that assumes frames of an identified window length to be independently and identically distributed (i.i.d.) -- which are then used to learn a recognition procedure. 
Owing to the non-constant sampling rates of data collected, varying duration and patterns of activities, the assumption of i.i.d.\ does not hold for data collected in smart homes.
Thus a straightforward approach to pre-process the data does not exist.

In this work, we address the aforementioned challenge, specifically through the identification of underlying structural constructs that make up activity sequences collected in the home. 
\textbf{Structural constructs are defined as the underlying unique sub-units or components that either constitute an activity or are relevant to it.}
For example, the activity of `Bed to Toilet' will be made of underlying constructs such as \textit{i)} wake up (sensor event in the bedroom triggers); \textit{ii)} use bathroom (sensor in bathroom); and, \textit{iii)} go back to bed / get ready for day (sensor event trigger elsewhere in home).
Existing smart home simulators e.g. VirtualHome \cite{puig2018virtualhome} use similar constructs (defined as categories) to generate data for a given activity, used to train robots assisting residents. 

We use Large language models (LLMs) to automatically identify these underlying constructs. 
With the advancements in LLMs, trained on internet-scale corpora, the consensus in the community is that these models rely on a semantic understanding of the underlying text to learn contextual mappings \cite{zhao2023survey, minaee2024large}.
Prior work \cite{hiremath2022bootstrapping} has shown the importance of capturing local context through learning an embedding using the BERT pre-trained model. 
In this work we extend this idea of using context, and scale it to include global information.

Utilizing the capabilities of LLMs to generate text, we construct coherent paragraphs corresponding to sensor event triggers, where individual input sentences include information corresponding to location and time. 
A summary is generated from the multiple sensor event triggers corresponding to a given activity.
In a ``game of LLMs'', we use one family of LLMs to generate these coherent paragraphs whereas another family of LLMs is then used to query for the underlying structural constructs. 

Through our explorations on the publicly available  CASAS benchmark datasets \cite{cook2012casas}, we aim to identify \textit{i)} which of the activities recorded in the homes are made of the hypothesized underlying constructs; and, \textit{ii)} categories of underlying constructs -- whether they are action-based or event-based. 
In the case of most activities of interest, constructs that are logical and relevant to the activity of interest are identified. 
The action-based constructs are listed in the sequence of occurrence that characterizes the corresponding activity. 
Practitioners can analyse the shortcomings of those activities where relevant constructs are not identified, to improve instrumentation of future homes.

\section{Related Work}
\label{sec2-related-work}
In the following section, we situate our proposed approach in related work that identifies  components of activity sequences known as sub-units, action units, events etc., to improve the performance of the activity recognition task. 
We aim to utilize the capabilities of large language models in identifying the aforementioned temporal constructs in the application scenario of smart homes. 

\subsection{Temporal Constructs in Sequential Data}
The activity recognition task is to identify activities of interest from continuous data streams.
Given the sequential nature of activities, approaches are required that model their temporal nature \cite{hiremath2021role, shao2024beyond}. 
Literature from the domains of hand-writing recognition, speech recognition and natural language processing, consider units of analysis that, when analyzed over longer context windows, are classified into relevant classes of interest \cite{plotz2009markov}. 
Similarly, in sensor based HAR activities comprise of units of analysis. 
Most work in sensor-based HAR, uses the same context length -- frames -- for both feature extraction and sequence modeling for activity classification \cite{bulling2014tutorial}. 
While this has shown to perform well through empirical research \cite{dehghani2019quantitative,bock2021improving,gupta2021deep}, the assumption that each of the units of analysis are independently and identically distributed may not necessarily hold. 
Statistical analysis has revealed that adjacent segments of time-series data, result in \textit{neighborhood bias}, and further influence performance estimates. 
This is observed due to adjacent frames of a given activity instance, which are highly similar appearing in both the train and test sets during the evualation of the activity recognition system \cite{hammerla2015let}.


As observed in gesture-based HAR systems, complex gestures are known to comprise of individual components \cite{mantyla2000hand}.
Once identified such components are modeled using sequential procedures like HMMs, LSTMs, or RNNs in general \cite{guan2017ensembles, ordonez2016deep}.
Previous work showcases the benefit of using smaller length segments to learn features and longer length sequences over an extended context to model activities \cite{hiremath2021role}.
More recently, in \cite{shao2024beyond} authors use attention from intra- and inter- frame attention to learn characteristics within individual frames and to capture longer contextual relationships across multiple frames.
These developed approaches showcase the benefits of identifying the underlying units and in capturing the temporal dynamics of activities to improve the performance of the recognition task. 
Discovering motifs in underlying sensor data, uses the Piece-wise Linear Aggregation (PAA) and Symbolic Aggregate approximation (SAX) to identify re-occurring units of signal data, which have been beneficial for both the activity recognition and discovery procedures \cite{lkhagva2006extended, yu2019novel, li2015piecewise}.  
In the realm of smart homes, actions units are learnt over analysis windows \cite{hiremath2022bootstrapping}, deemed to identify underlying movement patterns in the home. 
Motif models are learnt over these action units to identify \textit{relevant} sub-sequences of action units, that are used to perform activity recognition. 
More recently work in \cite{haresamudram2024towards} uses a vector quantization process via self-supervision to identify these underlying units of analysis \cite{makhoul1985vector, jaiswal2020survey, misra2020self}.

\section{Discovering Structural Concepts in Activities using LLMs}
\label{sec3:discovering-structure}
The key idea of the proposed analysis method is to identify underlying structural constructs that make up activity sequences, which can be further utilized to build an activity recognition system for smart homes.  
It comprises of designing prompts fed into one family of LLMs--GPT-4 (Open AI)--and obtaining constructs from a different family of LLMs -- Gemini (Google DeepMind). 
In the following sections we provide details about the LLMs used, the summarization procedure and the process of identifying the structural constructs. 

\begin{figure*}[t]
    \centering
    \includegraphics[width=\textwidth,height=46mm]{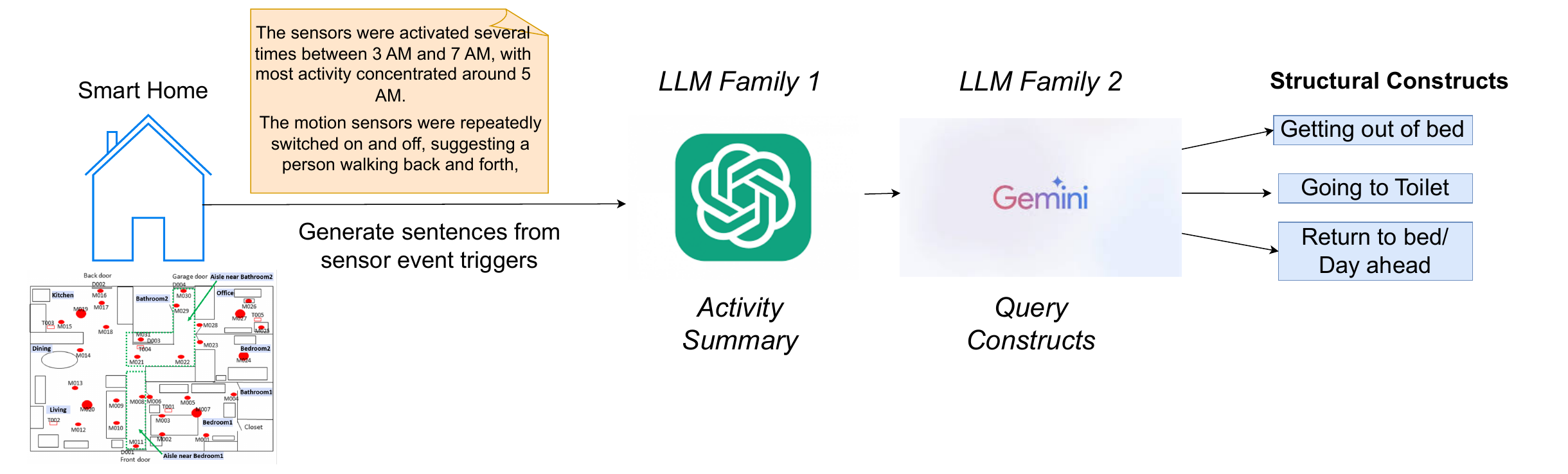}
    \caption{Overview of the proposed system. The proposed approach identifies the underlying structural concepts of activities observed in the smart home. First sentences detailing sensor event triggers are generated using information such as location and time of occurrence of activity \cite{thukral2024layout}. Next a family of LLMs (GPT-4) is used to obtain a summarized version of varied instances of these activities. Subsequently, another family of LLMs (Gemini) is queried to identify the structural constructs.}     
    \label{fig:discovery}
    \vspace*{-1.5em}
\end{figure*}

\subsection{Large Language Models}
We briefly discuss the large language models employed in this work below -- one for summarization (GPT-4) and the other for query (Gemini) to counter biases arising from training datasets used, model architectures optimized etc. \cite{carla2024large, rane2024gemini} for each of these LLMs.

\smallskip
\noindent
\textit{LLM Family 1 -- GPT-4:\,}
GPT-4 is a large language model developed by OpenAI, built on the foundation of the GPT (Generative Pre-trained Transformer) architecture \cite{achiam2023gpt,peng2023instruction,sanderson2023gpt}.
It's designed to engage in conversation, answer questions, generate text based on prompts, and perform various language-related tasks. 
The model learns from large datasets of text from the internet, allowing it to generate human-like responses across a wide range of topics.
GPT-4's architecture involves multiple layers of transformers, a type of neural network architecture that has been particularly successful in natural language processing (NLP) tasks. 
GPT-4 (and it's previous versions) are generative language models that can create content, offer explanations and provide assistance across diverse tasks and topics.
These transformers are trained on vast amounts of text data to learn patterns and relationships within language, enabling GPT-4 to generate coherent and contextually relevant responses.

\smallskip
\noindent
\textit{LLM Family 2 -- Gemini:\,}
Gemini references a family of Large Language Models (LLMs) which is the core technology behind several Google products \cite{mcintosh2023google, team2024gemma}. 
It's a powerful AI system capable of understanding and responding to different kinds of information, including text, code, audio, images, and video. 
Launched in December 2023, it succeeds previous models like LaMDA and PaLM 2. 
There are three variations:
\textit{i)} Gemini Ultra;
\textit{ii)} Gemini Pro; and, 
\textit{iii)} Gemini Nano.
We used Gemini Pro in this work. 

\subsection{GPT-4: Activity summarization}
To generate a summarized description of a given activity, the capabilities of GPT-4 are utilized.
Multiple paragraphs describing activity instances for a given activity are input to the LLM. 
On observing these descriptions the LLM is asked to produce a summary, where the idea is that this summary captures the important aspects of the given activity.
The prompt is as below:
\begin{quote}
\texttt{\noindent
       You are an AI assistant that is helping in generating a summary from diverse texts and adding a context to each sensor readings leveraging world knowledge
        }

\vspace*{0.2em}
\texttt{\noindent
        Please generate short summarized text (1) from the paragraphs of given activity descriptions.
        Ignore the temperature sensors. 
        Retain the time of occurrence of activity. 
        You will be given different paragraphs of the activity. }
        
\vspace*{0.5em}
 \texttt{\noindent        
        The input has format: (Paragraph: Text detailing sensor event triggers for given activity). The output should be a json (key: (activity)) containing the summarized paragraph.}
\end{quote}

Sensor event triggers for an activity instance are converted to sentences using the procedure in \cite{thukral2024layout}.
Each sensor event is mapped to a location and sensor type. 
The location for the sensor event trigger is obtained from annotations described in \cite{hiremath2022bootstrapping, thukral2024layout}.
Although we encode the temperature sensors while constructing these sentences, following previous work in \cite{casas_drop_temperature} we ask the LLM to ignore this information when generating the summary.
These sensors record changes in temperature observed in the smart home and do not capture movement.
The time of occurrence of an activity provides context information, shown to be an important indicator for some activities (e.g., Bed to Toilet) is retained while generating the summary.
Time between consecutive sensor event triggers is computed and added to the sentence. 

\subsection{Gemini: Identifying activity specific structural constructs}
Next, we use Gemini to query for the underlying constructs from the summarized activity text. 
We switch to a different family of LLMs to avoid bias arising from prompting and querying from the same LLM (GPT-4).
Model bias is categorized as: \textit{i)} embedding-based; \textit{ii)} probability-based; and, \textit{iii)} generated text-based \cite{gallegos2024bias}.
The summarized text for a given activity is provided as input, but the label of the activity is withheld.

\begin{quote}
   \texttt{\noindent
   You are an AI assistant helping with identifying categories of a summarized activity leveraging world knowledge."
   The summary of the given activity is (summary)."
   Can you provide the sub-actions that make up this activity?"
   }
\end{quote}
   
An example summarized activity text (from GPT-4) for an activity of of Meal\_Preparation (in CASAS-Aruba) is as follows -  {\fontfamily{pcr}\selectfont\small
`On multiple occasions, the motion sensor in the Kitchen and the area between the Kitchen and Dining area were triggered, indicating activity consistent with meal preparation. The sensors were activated at various \\ times throughout the day, with notable activity around 5:30 PM and 6:54 PM, as well as in the morning hours. The motion was detected as someone moved around the Kitchen, likely cooking or preparing food, and \\ occasionally moving to the Dining area, possibly to set the table or gather supplies. Temperature sensors, which are to be ignored, sporadically reported changes, but these do not directly correlate with the meal \\ preparation activity.' }

\section{Experimental Evaluation}
\label{sec4:experimental-evaluation}
We hypothesize that the discovery of underlying structural constructs in activity patterns is beneficial to  activity recognition. 
Through this work we identify and categorize these constructs and propose a method to combine these identified concepts in future work.
Technical foundations can be explored to develop the sequential modeling procedure, built on identified constructs.

\subsection{Datasets and Data Pre-Processing}
\label{sec:results-datasets}
\noindent

\begin{figure}[t]
    \vspace*{-1em}
    \subfigure[CASAS-Aruba]
    { \hspace*{0.8cm}  
    \centering
        \includegraphics[keepaspectratio, width=.5\textwidth, height=35mm]{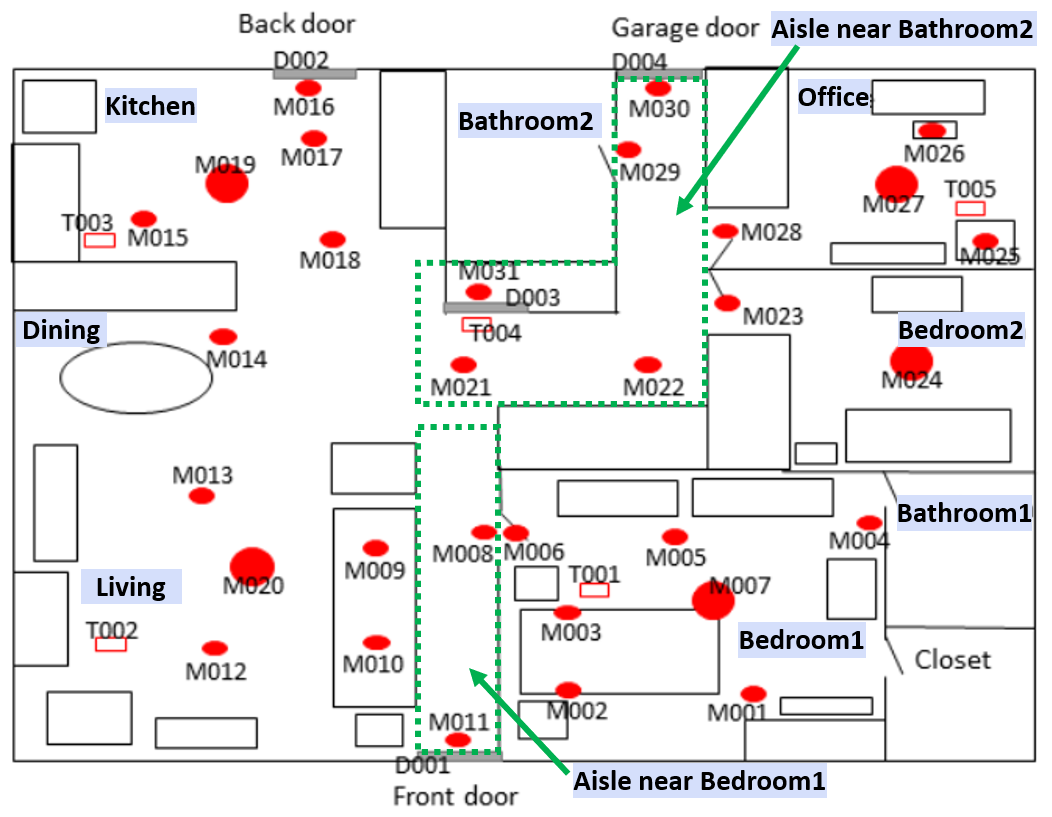}
        \label{fig:aruba}
    }
     \subfigure[CASAS-Milan]
    { 
     \hspace*{0.8cm}  
    \centering
        \includegraphics[keepaspectratio, width=.5\textwidth, height=35mm]{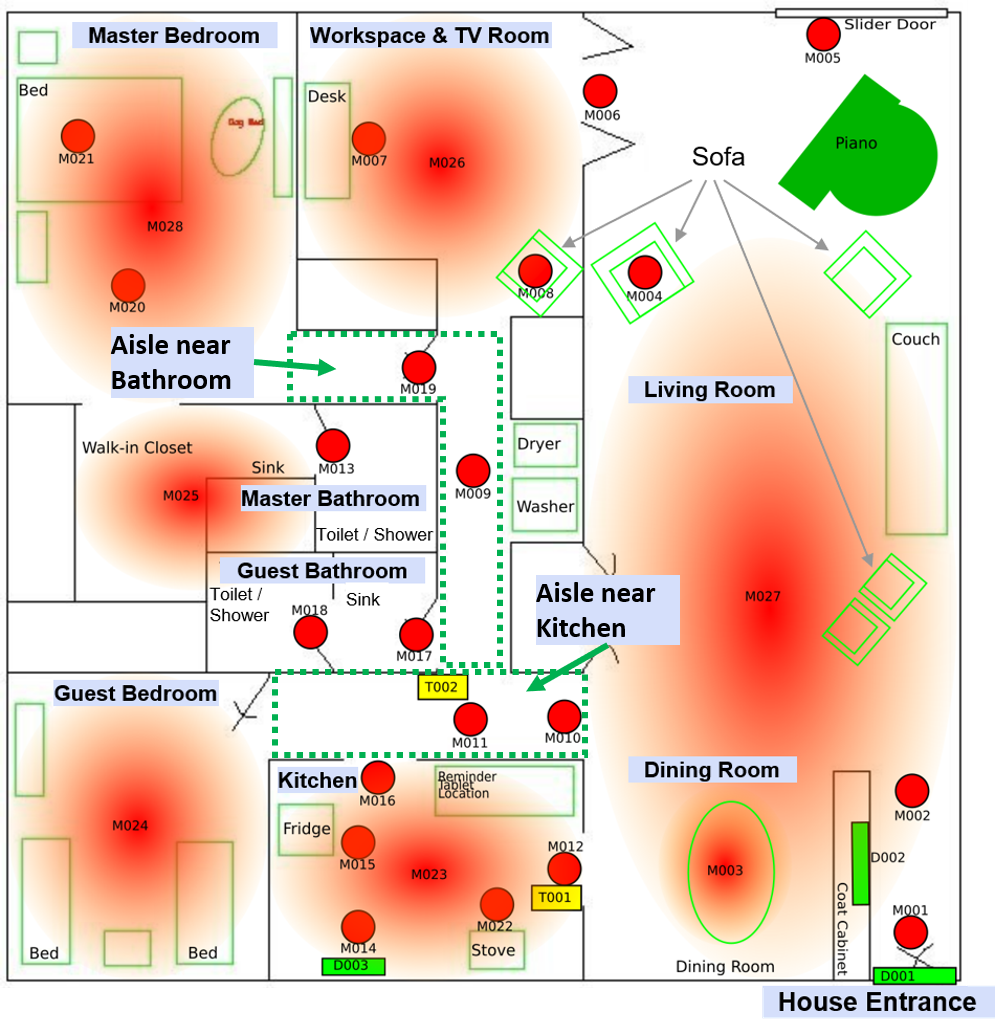}
        \label{fig:milan}
    }\\
    \caption{Floor plans for Smart Homes used for our experimental evaluation: (a) CASAS-Aruba and (b) CASAS-Milan  (taken with permission from \cite{cook2012casas}). 
    Annotations for locations are used with permission from \cite{hiremath2022bootstrapping}.}
     \label{fig:layout}
    \vspace*{-1em}
\end{figure}

Our explorations in this work are based on publicly available benchmarked datasets, collected as part of the Center of Advanced Studies in Adaptive Systems (CASAS), with ground truth annotations provided by residents \cite{cook2012casas}.
We specifically use the CASAS-Aruba and CASAS-Milan datasets, where data is collected over a span of 219 and 92 days, respectively.
Home layouts (with location specifications) of these datasets is shown in Fig. \ref{fig:layout}.
Both these are single-resident households, however, CASAS-Milan also houses a pet.
The number of activities analyzed to identify constructs are listed in Tab. \ref{tab:results-constructs:aruba} and \ref{tab:results-constructs:milan}.
Activities from the dataset have been combined according to previous work in \cite{liciotti2020sequential, thukral2024layout}. 
We ignore the `Other' (`null') class since this is representative of background data we are not interested in. 

Data in these homes is collected through ambient sensors -- Motion (denoted as `M\#\#\#'), Door (denoted as `D\#\#\#') and Temperature (denoted as `T\#\#\#'). 
The states of the Door sensors are either OPEN or CLOSE and those of the Motion sensors are either ON or OFF. 
Motion detection varies between localized areas and detection over a wider area, whereas door sensor triggers when using the various instrumented doors in the home. 
Temperature sensors record changes in the home, but do not capture movement. 
In order to generate summaries of activity sequences we use contextual information such as absolute timestamps and location information as in previous works \cite{hiremath2022bootstrapping, thukral2024layout}.

\subsection{Determining Structural Concepts}
\label{sec:results-concepts}
\noindent
The identification of structural constructs is a three step procedure -- \textit{i)} encode sensor event triggers as sentences; \textit{ii)} generate a summary of an activity by using \textit{`n'} varied sentences corresponding to the given activity (LLM1); and, \textit{iii)} use the summary to query for discovery of structural constructs (LLM2). 

A large number of such encoded sentences (where `n' is the maximum number of sentences that could be provided) are input to LLM1 for summarization.
The constructs identified are further classified into two categories: \textit{i)} event-based constructs; and, \textit{ii)} action-based constructs. 
Event-based constructs are representative of sub-actions that do not necessarily follow a sequence of occurrence.  
For example, in CASAS-Aruba, ``Relax" identified `Sitting' and `Walking' as sub-actions, which do not have to be in that particular order. 
Action-based constructs represent sub-actions that are commonly known to follow a sequence of occurrence. 
``Bed\_to\_Toilet" clearly identified sub-actions that must occur in the sequence they generated. 
The decision of category for the constructs is based on the description in the generated summary.
Identified constructs and their corresponding categories are listed in Tab. \ref{tab:results-constructs:aruba} 
and Tab. \ref{tab:results-constructs:milan} for the CASAS-Aruba and CASAS-Milan datasets, respectively.
Constructs in green depict those that are \textit{relevant} to the activity. 
Filtering and clarification procedures need to be employed for constructs not relevant to an activity (depicted in red).
An interesting observation is for `Take\_Medicine', where identified constructs are related to the activity of accessing and taking medicine, yet, isn't identified as a sub-action in itself (in the absence of specific contextual triggers -- such as opening kitchen / medicine cabinet).

Although, this work is aimed at delivering a proof-of-concept to identify constructs, it is not straightforward to validate them, given the absence of ground truth at this granularity. 
However, as in \cite{rashidi2009keeping}, a resident in the loop, through an active learning procedure can be asked to confirm both: \textit{i)} the \textit{number} of identified constructs; and, \textit{ii)} \textit{category} of the constructs determined. 
For the action-based constructs, the first identified sub-action provides a start-up trigger (and to possibly model time duration) -- thus, providing the necessary contextual information to identify an activity.
Previous work in \cite{rashidi2009keeping} required domain knowledge to do this, which is now possible with the use of LLMs.

\begin{table}[t]
    \centering
    \caption{Identified Structural Constructs for CASAS-Aruba. Color coding for the constructs is explained in text.}
    \label{tab:results-constructs:aruba}
    \small
    \begin{tabular}{p{1.4cm}|p{5.2cm}|p{0.9cm}}
        \textbf{Activity} & 
        \textbf{Identified Structural Constructs}& 
        \textbf{Type of constructs} \\
        \hline
        Sleeping & \positive{1. Sleeping(periods of no movement, interspersed with periods of movement. The temperature readings remain relatively stable throughout; 2. Movement (motion sensor activity;  moving around in the area between the living room and home entrance aisle); 3. Rest(associated with the periods of no movement)} & Event \\
        \hline
        Meal\_Prep &   \positive{1. Gathering ingredients; 2. Preparing ingredients (e.g., chopping, slicing, dicing); 3. Cooking ingredients (e.g., boiling, frying, baking); 4. Setting the table; 5. Serving the meal} & Action \\
         \hline
        Relax & \positive{1. Sitting; 2. Walking} & Event \\
         \hline
        Work & \positive{1. Movement(motion sensor is triggered frequently, indicating movement or presence in the area); 2. Sensor switching(suggesting continuous activity); 3. Work-related tasks(likely work-related, as it is taking place in an office space)} & Event  \\
        \hline
        Eating & \positive{1. Food preparation; 2.Cooking; 3. Eating; 4.Cleaning up} & Action\\
        \hline
        Bed\_to\_Toilet &  \positive{1. Getting out of bed; 2. Walking to the toilet; 3. Using the toilet
        4. Walking back to bed or starting the day} & Action \\
        \hline
        Enter\_Home & \positive{1. Open door; 2. Enter home; 3. Close door; 4.Move within entrance area} & Action \\
        \hline
        Leave\_Home & \positive{1. Open door; 2. Exit home; 3. Close door} & Action \\
        \hline
        Housekeeping & \negative{No summary generated from LLM1}  & -- \\
    \end{tabular}
\end{table}

\begin{table}[t]
    \centering
    \caption{Identified Structural Constructs for CASAS-Milan. Color coding for constructs is explained in text.}
    \label{tab:results-constructs:milan}
    \small
    \begin{tabular}{p{1.4cm}|p{5.2cm}|p{0.9cm}}
        \textbf{Activity} & 
        \textbf{Identified Structural Constructs}& 
        \textbf{Type of constructs} \\
        \hline
        Sleeping & \positive{1. Go to bed; 2. Sleep; 3. Waking up briefly; 4. Visiting the kitchen; 5. Visiting the bathroom} & Event  \\
        \hline
        Kitchen\_Act & \positive{1.Cooking (sensors near stove); 2. Refrigeration (sensor near fridge); 3. Movement (sensors near corridor, dining room, living room); 4. Door opening/closing (kitchen door sensor activation)}  & Event  \\
         \hline
        Guest\_Bath & \positive{1. Entering the guest bathroom; 2. Using the toilet; 3. Using the shower; 4. Washing hands; 5. Exiting the guest bathroom} & Action \\
         \hline
        Read & \negative{1. Motion in living room;  2. Temperature changes in the kitchen and corridor (to be ignored); 3. Movement in the dining room and near the home entrance; 4. Interactions with the slider door in the living room} & Event \\
        \hline
        Master\_Bath & \positive{1. Dressing; 2. Personal hygiene} & Event\\
        \hline
        Master\_Bed & \positive{1. Getting dressed; 2. Getting ready for the day; 3. Getting ready for bed; 4.Using the closet space} & Event \\
        \hline
        Watch\_TV & \positive{1. Sitting on couch; 2. Getting up to adjust TV; 3. Changing channels; 4. Talking to others; 5. Use restroom or get a drink} &  Event \\
        \hline
        Desk\_Act & \positive{1. Working; 2. Studying; 3. Using a computer} & Event  \\
        \hline
        Dining\_Act & \positive{1. Eating; 2. Socializing; 3. Cleaning} & Event\\
        \hline
        Leave\_Home & \positive{1. Leave home preparation (person moved around the entrance and living room, possibly gathering belongings or preparing to leave); 2. Open Door (opened entrance door); 3. Exit Home (sensor near entrance); 4. Close Door (Close entrance door)} & Action  \\
        \hline
        Take\_Med & \negative{1. Preparing Food; 2. Cleaning; Other kitchen tasks (getting food; groceries; meals)}  & Event
        \\
          \hline
        Meditate & \negative{1. Motion in quest bedroom; 2. Movement throughout the living room, dining room, and kitchen area; 3. Return to the guest bedroom} & Event  \\
         \hline
         Bed\_to\_Toilet  & \positive{1. Get out of bed; 2. Walk to the walk-in closet; 3. Walk to the bathroom; 4. Use the bathroom; 5. Walk back to the walk-in closet; 6. Walk back to bed} & Action  \\
    \end{tabular}
\end{table}

\subsection{Utility of Identified Constructs: Activity Recognition}
\label{sec:results-ar}
\noindent
In \cite{bettadapura2013augmenting}, authors define an activity as a finite sequence of events \textit{observable} through videos. 
Regular expressions are sampled from these events through a probabilistic approach to match to the activity patterns observed in the datasets. 
These expressions are used in the recognition procedure to identify activities and anomalies in a video based surveillance systems. 

Data collected in smart homes is hypothesized to be made of similar ``observable" events, yet these are not readily available. 
We use LLMs to identify these underlying events. 
We envision the use of regular expressions as in \cite{bettadapura2013augmenting} for activity recognition in smart homes.

\section{Discussion and Future Work}
\label{sec5:discussion}
In this work we provided a proof-of-concept that establishes the existence of underlying structural constructs in activity sequences. 
Prior work has been successful in identifying the most frequently occurring activities, and improvements over segmentation boundaries through a maintenance procedure for these identified activities \cite{hiremath2022bootstrapping}. 
However, recognition of the short-duration and infrequent activities is a challenge. 
Our results in this work successfully identify the events that make up these challenging activities. 
Through mapping discretized data streams \cite{haresamudram2024towards} and the constructs identified (this work), as described in Sec. \ref{sec:results-ar} (\cite{bettadapura2013augmenting}) identifying these activities can be possible.
Concepts identified through this work can be extended beyond the task of activity recognition and can be used for assessing routines in homes. 

The prompts provided to both LLMs -- for summarization and query -- are straightforward in the tasks they are expected to do. 
Improvements to these prompts will be explored in future work. 
For example, LLMs are known to perform well when fine-tuned for specific tasks with (input, output) pairs of data. 
LLMs have been known to provide illogical (or nonsensical) responses (\cite{king2024sasha, leng2024imugpt}), and thus filtering procedures are generally employed. 
We aim to investigate both the fine-tuning and filtering procedures.
Additionally, we summarized activity sequences to compute one activity summary. 
Due to restrictions on the number of queries to the API's, we generate queries on a subset of all instances for a given activity. 
Identifying a way to generate a comprehensive summary over all activity instances is to be explored.

\section{Conclusion}
\label{sec6:conclusion}
We have experimentally evaluated and confirmed our hypothesis that activity sequences are made up of structural constructs.
Our experimental procedures identify these constructs using two publicly available smart home datasets.
We used two families of LLMs to do this, wherein, the first family is used to summarize the sensor event triggers observed in the home, whereas the second family of LLMs is used to identify the structural constructs. 
Further, we propose construction of a recognition pipeline using the identified constructs. 


\bibliographystyle{ACM-Reference-Format}
\bibliography{refs.bib}


\begin{thebibliography}{50}


\ifx \showCODEN    \undefined \def \showCODEN     #1{\unskip}     \fi
\ifx \showDOI      \undefined \def \showDOI       #1{#1}\fi
\ifx \showISBNx    \undefined \def \showISBNx     #1{\unskip}     \fi
\ifx \showISBNxiii \undefined \def \showISBNxiii  #1{\unskip}     \fi
\ifx \showISSN     \undefined \def \showISSN      #1{\unskip}     \fi
\ifx \showLCCN     \undefined \def \showLCCN      #1{\unskip}     \fi
\ifx \shownote     \undefined \def \shownote      #1{#1}          \fi
\ifx \showarticletitle \undefined \def \showarticletitle #1{#1}   \fi
\ifx \showURL      \undefined \def \showURL       {\relax}        \fi
\providecommand\bibfield[2]{#2}
\providecommand\bibinfo[2]{#2}
\providecommand\natexlab[1]{#1}
\providecommand\showeprint[2][]{arXiv:#2}

\bibitem[Achiam et~al\mbox{.}(2023)]%
        {achiam2023gpt}
\bibfield{author}{\bibinfo{person}{Josh Achiam}, \bibinfo{person}{Steven Adler}, \bibinfo{person}{Sandhini Agarwal}, \bibinfo{person}{Lama Ahmad}, \bibinfo{person}{Ilge Akkaya}, \bibinfo{person}{Florencia~Leoni Aleman}, \bibinfo{person}{Diogo Almeida}, \bibinfo{person}{Janko Altenschmidt}, \bibinfo{person}{Sam Altman}, \bibinfo{person}{Shyamal Anadkat}, {et~al\mbox{.}}} \bibinfo{year}{2023}\natexlab{}.
\newblock \showarticletitle{Gpt-4 technical report}.
\newblock \bibinfo{journal}{\emph{arXiv preprint arXiv:2303.08774}} (\bibinfo{year}{2023}).
\newblock


\bibitem[Asadzadeh et~al\mbox{.}(2022)]%
        {asadzadeh2022review}
\bibfield{author}{\bibinfo{person}{Mohammad Asadzadeh}, \bibinfo{person}{Ali Maher}, \bibinfo{person}{Mehrnoosh Jafari}, \bibinfo{person}{Khalil~A Mohammadzadeh}, {and} \bibinfo{person}{Seyed~Mojtaba Hosseini}.} \bibinfo{year}{2022}\natexlab{}.
\newblock \showarticletitle{A review study of the providing elderly care services in different countries}.
\newblock \bibinfo{journal}{\emph{Journal of Family Medicine and Primary Care}} \bibinfo{volume}{11}, \bibinfo{number}{2} (\bibinfo{year}{2022}), \bibinfo{pages}{458--465}.
\newblock


\bibitem[Belley et~al\mbox{.}(2014)]%
        {belley2014efficient}
\bibfield{author}{\bibinfo{person}{Corinne Belley}, \bibinfo{person}{Sebastien Gaboury}, \bibinfo{person}{Bruno Bouchard}, {and} \bibinfo{person}{Abdenour Bouzouane}.} \bibinfo{year}{2014}\natexlab{}.
\newblock \showarticletitle{An efficient and inexpensive method for activity recognition within a smart home based on load signatures of appliances}.
\newblock \bibinfo{journal}{\emph{Pervasive and Mobile Computing}}  \bibinfo{volume}{12} (\bibinfo{year}{2014}), \bibinfo{pages}{58--78}.
\newblock


\bibitem[Bettadapura et~al\mbox{.}(2013)]%
        {bettadapura2013augmenting}
\bibfield{author}{\bibinfo{person}{Vinay Bettadapura}, \bibinfo{person}{Grant Schindler}, \bibinfo{person}{Thomas Pl{\"o}tz}, {and} \bibinfo{person}{Irfan Essa}.} \bibinfo{year}{2013}\natexlab{}.
\newblock \showarticletitle{Augmenting bag-of-words: Data-driven discovery of temporal and structural information for activity recognition}. In \bibinfo{booktitle}{\emph{Proceedings of the IEEE Conference on Computer Vision and Pattern Recognition}}. \bibinfo{pages}{2619--2626}.
\newblock


\bibitem[Bock et~al\mbox{.}(2021)]%
        {bock2021improving}
\bibfield{author}{\bibinfo{person}{Marius Bock}, \bibinfo{person}{Alexander H{\"o}lzemann}, \bibinfo{person}{Michael Moeller}, {and} \bibinfo{person}{Kristof Van~Laerhoven}.} \bibinfo{year}{2021}\natexlab{}.
\newblock \showarticletitle{Improving deep learning for HAR with shallow LSTMs}. In \bibinfo{booktitle}{\emph{Proceedings of the 2021 ACM International Symposium on Wearable Computers}}. \bibinfo{pages}{7--12}.
\newblock


\bibitem[Bouchabou et~al\mbox{.}(2021)]%
        {bouchabou2021survey}
\bibfield{author}{\bibinfo{person}{Damien Bouchabou}, \bibinfo{person}{Sao~Mai Nguyen}, \bibinfo{person}{Christophe Lohr}, \bibinfo{person}{Benoit LeDuc}, {and} \bibinfo{person}{Ioannis Kanellos}.} \bibinfo{year}{2021}\natexlab{}.
\newblock \showarticletitle{A survey of human activity recognition in smart homes based on IoT sensors algorithms: Taxonomies, challenges, and opportunities with deep learning}.
\newblock \bibinfo{journal}{\emph{Sensors}} \bibinfo{volume}{21}, \bibinfo{number}{18} (\bibinfo{year}{2021}), \bibinfo{pages}{6037}.
\newblock


\bibitem[Bulling et~al\mbox{.}(2014)]%
        {bulling2014tutorial}
\bibfield{author}{\bibinfo{person}{Andreas Bulling}, \bibinfo{person}{Ulf Blanke}, {and} \bibinfo{person}{Bernt Schiele}.} \bibinfo{year}{2014}\natexlab{}.
\newblock \showarticletitle{A tutorial on human activity recognition using body-worn inertial sensors}.
\newblock \bibinfo{journal}{\emph{ACM Computing Surveys (CSUR)}} \bibinfo{volume}{46}, \bibinfo{number}{3} (\bibinfo{year}{2014}), \bibinfo{pages}{1--33}.
\newblock


\bibitem[Burke(2017)]%
        {burke2017sandwich}
\bibfield{author}{\bibinfo{person}{Ronald~J Burke}.} \bibinfo{year}{2017}\natexlab{}.
\newblock \showarticletitle{The sandwich generation: individual, family, organizational and societal challenges and opportunities}.
\newblock \bibinfo{journal}{\emph{The sandwich generation}} (\bibinfo{year}{2017}), \bibinfo{pages}{3--39}.
\newblock


\bibitem[Carl{\`a} et~al\mbox{.}(2024)]%
        {carla2024large}
\bibfield{author}{\bibinfo{person}{Matteo~Mario Carl{\`a}}, \bibinfo{person}{Gloria Gambini}, \bibinfo{person}{Antonio Baldascino}, \bibinfo{person}{Francesco Boselli}, \bibinfo{person}{Federico Giannuzzi}, \bibinfo{person}{Fabio Margollicci}, {and} \bibinfo{person}{Stanislao Rizzo}.} \bibinfo{year}{2024}\natexlab{}.
\newblock \showarticletitle{Large language models as assistance for glaucoma surgical cases: a ChatGPT vs. Google Gemini comparison}.
\newblock \bibinfo{journal}{\emph{Graefe's Archive for Clinical and Experimental Ophthalmology}} (\bibinfo{year}{2024}), \bibinfo{pages}{1--15}.
\newblock


\bibitem[Cook et~al\mbox{.}(2012)]%
        {cook2012casas}
\bibfield{author}{\bibinfo{person}{Diane~J Cook}, \bibinfo{person}{Aaron~S Crandall}, \bibinfo{person}{Brian~L Thomas}, {and} \bibinfo{person}{Narayanan~C Krishnan}.} \bibinfo{year}{2012}\natexlab{}.
\newblock \showarticletitle{CASAS: A smart home in a box}.
\newblock \bibinfo{journal}{\emph{Computer}} \bibinfo{volume}{46}, \bibinfo{number}{7} (\bibinfo{year}{2012}), \bibinfo{pages}{62--69}.
\newblock


\bibitem[Crandall and Cook(2012)]%
        {crandall2012smart}
\bibfield{author}{\bibinfo{person}{Aaron~S Crandall} {and} \bibinfo{person}{Diane~J Cook}.} \bibinfo{year}{2012}\natexlab{}.
\newblock \showarticletitle{Smart Home in a Box: A Large Scale Smart Home Deployment.}. In \bibinfo{booktitle}{\emph{Intelligent Environments (Workshops)}}. \bibinfo{pages}{169--178}.
\newblock


\bibitem[Dehghani et~al\mbox{.}(2019)]%
        {dehghani2019quantitative}
\bibfield{author}{\bibinfo{person}{Akbar Dehghani}, \bibinfo{person}{Omid Sarbishei}, \bibinfo{person}{Tristan Glatard}, {and} \bibinfo{person}{Emad Shihab}.} \bibinfo{year}{2019}\natexlab{}.
\newblock \showarticletitle{A quantitative comparison of overlapping and non-overlapping sliding windows for human activity recognition using inertial sensors}.
\newblock \bibinfo{journal}{\emph{Sensors}} \bibinfo{volume}{19}, \bibinfo{number}{22} (\bibinfo{year}{2019}), \bibinfo{pages}{5026}.
\newblock


\bibitem[Deldari et~al\mbox{.}(2022)]%
        {deldari2022beyond}
\bibfield{author}{\bibinfo{person}{Shohreh Deldari}, \bibinfo{person}{Hao Xue}, \bibinfo{person}{Aaqib Saeed}, \bibinfo{person}{Jiayuan He}, \bibinfo{person}{Daniel~V Smith}, {and} \bibinfo{person}{Flora~D Salim}.} \bibinfo{year}{2022}\natexlab{}.
\newblock \showarticletitle{Beyond just vision: A review on self-supervised representation learning on multimodal and temporal data}.
\newblock \bibinfo{journal}{\emph{arXiv preprint arXiv:2206.02353}} (\bibinfo{year}{2022}).
\newblock


\bibitem[Gallegos et~al\mbox{.}(2024)]%
        {gallegos2024bias}
\bibfield{author}{\bibinfo{person}{Isabel~O Gallegos}, \bibinfo{person}{Ryan~A Rossi}, \bibinfo{person}{Joe Barrow}, \bibinfo{person}{Md~Mehrab Tanjim}, \bibinfo{person}{Sungchul Kim}, \bibinfo{person}{Franck Dernoncourt}, \bibinfo{person}{Tong Yu}, \bibinfo{person}{Ruiyi Zhang}, {and} \bibinfo{person}{Nesreen~K Ahmed}.} \bibinfo{year}{2024}\natexlab{}.
\newblock \showarticletitle{Bias and fairness in large language models: A survey}.
\newblock \bibinfo{journal}{\emph{Computational Linguistics}} (\bibinfo{year}{2024}), \bibinfo{pages}{1--79}.
\newblock


\bibitem[Gazis and Katsiri(2021)]%
        {gazis2021smart}
\bibfield{author}{\bibinfo{person}{Alexandros Gazis} {and} \bibinfo{person}{Eleftheria Katsiri}.} \bibinfo{year}{2021}\natexlab{}.
\newblock \showarticletitle{Smart home IoT sensors: Principles and applications a review of low-cost and low-power solutions}.
\newblock \bibinfo{journal}{\emph{International Journal on Engineering Technologies and Informatics}} \bibinfo{volume}{2}, \bibinfo{number}{1} (\bibinfo{year}{2021}), \bibinfo{pages}{19--23}.
\newblock


\bibitem[Guan and Pl{\"o}tz(2017)]%
        {guan2017ensembles}
\bibfield{author}{\bibinfo{person}{Yu Guan} {and} \bibinfo{person}{Thomas Pl{\"o}tz}.} \bibinfo{year}{2017}\natexlab{}.
\newblock \showarticletitle{Ensembles of deep lstm learners for activity recognition using wearables}.
\newblock \bibinfo{journal}{\emph{Proceedings of the ACM on interactive, mobile, wearable and ubiquitous technologies}} \bibinfo{volume}{1}, \bibinfo{number}{2} (\bibinfo{year}{2017}), \bibinfo{pages}{1--28}.
\newblock


\bibitem[Gupta(2021)]%
        {gupta2021deep}
\bibfield{author}{\bibinfo{person}{Saurabh Gupta}.} \bibinfo{year}{2021}\natexlab{}.
\newblock \showarticletitle{Deep learning based human activity recognition (HAR) using wearable sensor data}.
\newblock \bibinfo{journal}{\emph{International Journal of Information Management Data Insights}} \bibinfo{volume}{1}, \bibinfo{number}{2} (\bibinfo{year}{2021}), \bibinfo{pages}{100046}.
\newblock


\bibitem[Hammerla and Pl{\"o}tz(2015)]%
        {hammerla2015let}
\bibfield{author}{\bibinfo{person}{Nils~Y Hammerla} {and} \bibinfo{person}{Thomas Pl{\"o}tz}.} \bibinfo{year}{2015}\natexlab{}.
\newblock \showarticletitle{Let's (not) stick together: pairwise similarity biases cross-validation in activity recognition}. In \bibinfo{booktitle}{\emph{Proceedings of the 2015 ACM international joint conference on pervasive and ubiquitous computing}}. \bibinfo{pages}{1041--1051}.
\newblock


\bibitem[Haresamudram et~al\mbox{.}(2024)]%
        {haresamudram2024towards}
\bibfield{author}{\bibinfo{person}{Harish Haresamudram}, \bibinfo{person}{Irfan Essa}, {and} \bibinfo{person}{Thomas Ploetz}.} \bibinfo{year}{2024}\natexlab{}.
\newblock \showarticletitle{Towards Learning Discrete Representations via Self-Supervision for Wearables-Based Human Activity Recognition}.
\newblock \bibinfo{journal}{\emph{Sensors}} \bibinfo{volume}{24}, \bibinfo{number}{4} (\bibinfo{year}{2024}), \bibinfo{pages}{1238}.
\newblock


\bibitem[Hiremath et~al\mbox{.}(2022)]%
        {hiremath2022bootstrapping}
\bibfield{author}{\bibinfo{person}{Shruthi~K Hiremath}, \bibinfo{person}{Yasutaka Nishimura}, \bibinfo{person}{Sonia Chernova}, {and} \bibinfo{person}{Thomas Pl{\"o}tz}.} \bibinfo{year}{2022}\natexlab{}.
\newblock \showarticletitle{Bootstrapping human activity recognition systems for smart homes from scratch}.
\newblock \bibinfo{journal}{\emph{Proceedings of the ACM on Interactive, Mobile, Wearable and Ubiquitous Technologies}} \bibinfo{volume}{6}, \bibinfo{number}{3} (\bibinfo{year}{2022}), \bibinfo{pages}{1--27}.
\newblock


\bibitem[Hiremath and Ploetz(2021)]%
        {hiremath2021role}
\bibfield{author}{\bibinfo{person}{Shruthi~Kashinath Hiremath} {and} \bibinfo{person}{Thomas Ploetz}.} \bibinfo{year}{2021}\natexlab{}.
\newblock \showarticletitle{On the role of context length for feature extraction and sequence modeling in human activity recognition}. In \bibinfo{booktitle}{\emph{Proceedings of the 2021 ACM International Symposium on Wearable Computers}}. \bibinfo{pages}{13--17}.
\newblock


\bibitem[Jaiswal et~al\mbox{.}(2020)]%
        {jaiswal2020survey}
\bibfield{author}{\bibinfo{person}{Ashish Jaiswal}, \bibinfo{person}{Ashwin~Ramesh Babu}, \bibinfo{person}{Mohammad~Zaki Zadeh}, \bibinfo{person}{Debapriya Banerjee}, {and} \bibinfo{person}{Fillia Makedon}.} \bibinfo{year}{2020}\natexlab{}.
\newblock \showarticletitle{A survey on contrastive self-supervised learning}.
\newblock \bibinfo{journal}{\emph{Technologies}} \bibinfo{volume}{9}, \bibinfo{number}{1} (\bibinfo{year}{2020}), \bibinfo{pages}{2}.
\newblock


\bibitem[King et~al\mbox{.}(2024)]%
        {king2024sasha}
\bibfield{author}{\bibinfo{person}{Evan King}, \bibinfo{person}{Haoxiang Yu}, \bibinfo{person}{Sangsu Lee}, {and} \bibinfo{person}{Christine Julien}.} \bibinfo{year}{2024}\natexlab{}.
\newblock \showarticletitle{Sasha: creative goal-oriented reasoning in smart homes with large language models}.
\newblock \bibinfo{journal}{\emph{Proceedings of the ACM on Interactive, Mobile, Wearable and Ubiquitous Technologies}} \bibinfo{volume}{8}, \bibinfo{number}{1} (\bibinfo{year}{2024}), \bibinfo{pages}{1--38}.
\newblock


\bibitem[Leng et~al\mbox{.}(2024)]%
        {leng2024imugpt}
\bibfield{author}{\bibinfo{person}{Zikang Leng}, \bibinfo{person}{Amitrajit Bhattacharjee}, \bibinfo{person}{Hrudhai Rajasekhar}, \bibinfo{person}{Lizhe Zhang}, \bibinfo{person}{Elizabeth Bruda}, \bibinfo{person}{Hyeokhyen Kwon}, {and} \bibinfo{person}{Thomas Pl{\"o}tz}.} \bibinfo{year}{2024}\natexlab{}.
\newblock \showarticletitle{IMUGPT 2.0: Language-Based Cross Modality Transfer for Sensor-Based Human Activity Recognition}.
\newblock \bibinfo{journal}{\emph{arXiv preprint arXiv:2402.01049}} (\bibinfo{year}{2024}).
\newblock


\bibitem[Li(2015)]%
        {li2015piecewise}
\bibfield{author}{\bibinfo{person}{Hailin Li}.} \bibinfo{year}{2015}\natexlab{}.
\newblock \showarticletitle{Piecewise aggregate representations and lower-bound distance functions for multivariate time series}.
\newblock \bibinfo{journal}{\emph{Physica A: Statistical Mechanics and its Applications}}  \bibinfo{volume}{427} (\bibinfo{year}{2015}), \bibinfo{pages}{10--25}.
\newblock


\bibitem[Liciotti et~al\mbox{.}(2020)]%
        {liciotti2020sequential}
\bibfield{author}{\bibinfo{person}{Daniele Liciotti}, \bibinfo{person}{Michele Bernardini}, \bibinfo{person}{Luca Romeo}, {and} \bibinfo{person}{Emanuele Frontoni}.} \bibinfo{year}{2020}\natexlab{}.
\newblock \showarticletitle{A sequential deep learning application for recognising human activities in smart homes}.
\newblock \bibinfo{journal}{\emph{Neurocomputing}}  \bibinfo{volume}{396} (\bibinfo{year}{2020}), \bibinfo{pages}{501--513}.
\newblock


\bibitem[Lkhagva et~al\mbox{.}(2006)]%
        {lkhagva2006extended}
\bibfield{author}{\bibinfo{person}{Battuguldur Lkhagva}, \bibinfo{person}{Yu Suzuki}, {and} \bibinfo{person}{Kyoji Kawagoe}.} \bibinfo{year}{2006}\natexlab{}.
\newblock \showarticletitle{Extended SAX: Extension of symbolic aggregate approximation for financial time series data representation}.
\newblock \bibinfo{journal}{\emph{DEWS2006 4A-i8}}  \bibinfo{volume}{7} (\bibinfo{year}{2006}).
\newblock


\bibitem[Majumder et~al\mbox{.}(2017)]%
        {majumder2017smart}
\bibfield{author}{\bibinfo{person}{Sumit Majumder}, \bibinfo{person}{Emad Aghayi}, \bibinfo{person}{Moein Noferesti}, \bibinfo{person}{Hamidreza Memarzadeh-Tehran}, \bibinfo{person}{Tapas Mondal}, \bibinfo{person}{Zhibo Pang}, {and} \bibinfo{person}{M~Jamal Deen}.} \bibinfo{year}{2017}\natexlab{}.
\newblock \showarticletitle{Smart homes for elderly healthcare—Recent advances and research challenges}.
\newblock \bibinfo{journal}{\emph{Sensors}} \bibinfo{volume}{17}, \bibinfo{number}{11} (\bibinfo{year}{2017}), \bibinfo{pages}{2496}.
\newblock


\bibitem[Makhoul et~al\mbox{.}(1985)]%
        {makhoul1985vector}
\bibfield{author}{\bibinfo{person}{John Makhoul}, \bibinfo{person}{Salim Roucos}, {and} \bibinfo{person}{Herbert Gish}.} \bibinfo{year}{1985}\natexlab{}.
\newblock \showarticletitle{Vector quantization in speech coding}.
\newblock \bibinfo{journal}{\emph{Proc. IEEE}} \bibinfo{volume}{73}, \bibinfo{number}{11} (\bibinfo{year}{1985}), \bibinfo{pages}{1551--1588}.
\newblock


\bibitem[Mantyla et~al\mbox{.}(2000)]%
        {mantyla2000hand}
\bibfield{author}{\bibinfo{person}{V-M Mantyla}, \bibinfo{person}{J Mantyjarvi}, \bibinfo{person}{T Seppanen}, {and} \bibinfo{person}{Esa Tuulari}.} \bibinfo{year}{2000}\natexlab{}.
\newblock \showarticletitle{Hand gesture recognition of a mobile device user}. In \bibinfo{booktitle}{\emph{2000 IEEE International Conference on Multimedia and Expo. ICME2000. Proceedings. Latest Advances in the Fast Changing World of Multimedia (Cat. No. 00TH8532)}}, Vol.~\bibinfo{volume}{1}. IEEE, \bibinfo{pages}{281--284}.
\newblock


\bibitem[McIntosh et~al\mbox{.}(2023)]%
        {mcintosh2023google}
\bibfield{author}{\bibinfo{person}{Timothy~R McIntosh}, \bibinfo{person}{Teo Susnjak}, \bibinfo{person}{Tong Liu}, \bibinfo{person}{Paul Watters}, {and} \bibinfo{person}{Malka~N Halgamuge}.} \bibinfo{year}{2023}\natexlab{}.
\newblock \showarticletitle{From google gemini to openai q*(q-star): A survey of reshaping the generative artificial intelligence (ai) research landscape}.
\newblock \bibinfo{journal}{\emph{arXiv preprint arXiv:2312.10868}} (\bibinfo{year}{2023}).
\newblock


\bibitem[Minaee et~al\mbox{.}(2024)]%
        {minaee2024large}
\bibfield{author}{\bibinfo{person}{Shervin Minaee}, \bibinfo{person}{Tomas Mikolov}, \bibinfo{person}{Narjes Nikzad}, \bibinfo{person}{Meysam Chenaghlu}, \bibinfo{person}{Richard Socher}, \bibinfo{person}{Xavier Amatriain}, {and} \bibinfo{person}{Jianfeng Gao}.} \bibinfo{year}{2024}\natexlab{}.
\newblock \showarticletitle{Large language models: A survey}.
\newblock \bibinfo{journal}{\emph{arXiv preprint arXiv:2402.06196}} (\bibinfo{year}{2024}).
\newblock


\bibitem[Misra and Maaten(2020)]%
        {misra2020self}
\bibfield{author}{\bibinfo{person}{Ishan Misra} {and} \bibinfo{person}{Laurens van~der Maaten}.} \bibinfo{year}{2020}\natexlab{}.
\newblock \showarticletitle{Self-supervised learning of pretext-invariant representations}. In \bibinfo{booktitle}{\emph{Proceedings of the IEEE/CVF conference on computer vision and pattern recognition}}. \bibinfo{pages}{6707--6717}.
\newblock


\bibitem[Ord{\'o}{\~n}ez and Roggen(2016)]%
        {ordonez2016deep}
\bibfield{author}{\bibinfo{person}{Francisco~Javier Ord{\'o}{\~n}ez} {and} \bibinfo{person}{Daniel Roggen}.} \bibinfo{year}{2016}\natexlab{}.
\newblock \showarticletitle{Deep convolutional and lstm recurrent neural networks for multimodal wearable activity recognition}.
\newblock \bibinfo{journal}{\emph{Sensors}} \bibinfo{volume}{16}, \bibinfo{number}{1} (\bibinfo{year}{2016}), \bibinfo{pages}{115}.
\newblock


\bibitem[Pal et~al\mbox{.}(2018)]%
        {pal2018internet}
\bibfield{author}{\bibinfo{person}{Debajyoti Pal}, \bibinfo{person}{Suree Funilkul}, \bibinfo{person}{Nipon Charoenkitkarn}, {and} \bibinfo{person}{Prasert Kanthamanon}.} \bibinfo{year}{2018}\natexlab{}.
\newblock \showarticletitle{Internet-of-things and smart homes for elderly healthcare: An end user perspective}.
\newblock \bibinfo{journal}{\emph{IEEE Access}}  \bibinfo{volume}{6} (\bibinfo{year}{2018}), \bibinfo{pages}{10483--10496}.
\newblock


\bibitem[Pashazade et~al\mbox{.}(2023)]%
        {pashazade2023explaining}
\bibfield{author}{\bibinfo{person}{Hakime Pashazade}, \bibinfo{person}{Masoomeh Maarefvand}, \bibinfo{person}{Kianoush Abdi}, {and} \bibinfo{person}{Yadollah~Abolfathi Momtaz}.} \bibinfo{year}{2023}\natexlab{}.
\newblock \showarticletitle{Explaining the Process of Caregiving by Sandwich Generation}.
\newblock  (\bibinfo{year}{2023}).
\newblock


\bibitem[Peng et~al\mbox{.}(2023)]%
        {peng2023instruction}
\bibfield{author}{\bibinfo{person}{Baolin Peng}, \bibinfo{person}{Chunyuan Li}, \bibinfo{person}{Pengcheng He}, \bibinfo{person}{Michel Galley}, {and} \bibinfo{person}{Jianfeng Gao}.} \bibinfo{year}{2023}\natexlab{}.
\newblock \showarticletitle{Instruction tuning with gpt-4}.
\newblock \bibinfo{journal}{\emph{arXiv preprint arXiv:2304.03277}} (\bibinfo{year}{2023}).
\newblock


\bibitem[Pl{\"O}tz(2021)]%
        {plotz2021applying}
\bibfield{author}{\bibinfo{person}{Thomas Pl{\"O}tz}.} \bibinfo{year}{2021}\natexlab{}.
\newblock \showarticletitle{Applying machine learning for sensor data analysis in interactive systems: Common pitfalls of pragmatic use and ways to avoid them}.
\newblock \bibinfo{journal}{\emph{ACM Computing Surveys (CSUR)}} \bibinfo{volume}{54}, \bibinfo{number}{6} (\bibinfo{year}{2021}), \bibinfo{pages}{1--25}.
\newblock


\bibitem[Pl{\"o}tz and Fink(2009)]%
        {plotz2009markov}
\bibfield{author}{\bibinfo{person}{Thomas Pl{\"o}tz} {and} \bibinfo{person}{Gernot~A Fink}.} \bibinfo{year}{2009}\natexlab{}.
\newblock \showarticletitle{Markov models for offline handwriting recognition: a survey}.
\newblock \bibinfo{journal}{\emph{International Journal on Document Analysis and Recognition (IJDAR)}}  \bibinfo{volume}{12} (\bibinfo{year}{2009}), \bibinfo{pages}{269--298}.
\newblock


\bibitem[Prankit et~al\mbox{.}(2020)]%
        {casas_drop_temperature}
\bibfield{author}{\bibinfo{person}{Gupta Prankit}, \bibinfo{person}{Richard McClatchey}, {and} \bibinfo{person}{Praminda Caleb-Solly}.} \bibinfo{year}{2020}\natexlab{}.
\newblock \showarticletitle{Tracking changes in user activity from unlabelled smart home sensor data using unsupervised learning methods}. In \bibinfo{booktitle}{\emph{Neural Computing and Applications}}, Vol.~\bibinfo{volume}{32}. Springer, \bibinfo{pages}{12351--12362}.
\newblock


\bibitem[Puig et~al\mbox{.}(2018)]%
        {puig2018virtualhome}
\bibfield{author}{\bibinfo{person}{Xavier Puig}, \bibinfo{person}{Kevin Ra}, \bibinfo{person}{Marko Boben}, \bibinfo{person}{Jiaman Li}, \bibinfo{person}{Tingwu Wang}, \bibinfo{person}{Sanja Fidler}, {and} \bibinfo{person}{Antonio Torralba}.} \bibinfo{year}{2018}\natexlab{}.
\newblock \showarticletitle{Virtualhome: Simulating household activities via programs}. In \bibinfo{booktitle}{\emph{Proceedings of the IEEE conference on computer vision and pattern recognition}}. \bibinfo{pages}{8494--8502}.
\newblock


\bibitem[Rane et~al\mbox{.}(2024)]%
        {rane2024gemini}
\bibfield{author}{\bibinfo{person}{Nitin Rane}, \bibinfo{person}{Saurabh Choudhary}, {and} \bibinfo{person}{Jayesh Rane}.} \bibinfo{year}{2024}\natexlab{}.
\newblock \showarticletitle{Gemini Versus ChatGPT: Applications, Performance, Architecture, Capabilities, and Implementation}.
\newblock \bibinfo{journal}{\emph{Performance, Architecture, Capabilities, and Implementation (February 13, 2024)}} (\bibinfo{year}{2024}).
\newblock


\bibitem[Rashidi and Cook(2009)]%
        {rashidi2009keeping}
\bibfield{author}{\bibinfo{person}{Parisa Rashidi} {and} \bibinfo{person}{Diane~J Cook}.} \bibinfo{year}{2009}\natexlab{}.
\newblock \showarticletitle{Keeping the resident in the loop: Adapting the smart home to the user}.
\newblock \bibinfo{journal}{\emph{IEEE Transactions on systems, man, and cybernetics-part A: systems and humans}} \bibinfo{volume}{39}, \bibinfo{number}{5} (\bibinfo{year}{2009}), \bibinfo{pages}{949--959}.
\newblock


\bibitem[Sanderson(2023)]%
        {sanderson2023gpt}
\bibfield{author}{\bibinfo{person}{Katharine Sanderson}.} \bibinfo{year}{2023}\natexlab{}.
\newblock \showarticletitle{GPT-4 is here: what scientists think}.
\newblock \bibinfo{journal}{\emph{Nature}} \bibinfo{volume}{615}, \bibinfo{number}{7954} (\bibinfo{year}{2023}), \bibinfo{pages}{773}.
\newblock


\bibitem[Shao et~al\mbox{.}(2024)]%
        {shao2024beyond}
\bibfield{author}{\bibinfo{person}{Shuai Shao}, \bibinfo{person}{Yu Guan}, {and} \bibinfo{person}{Victor Sanchez}.} \bibinfo{year}{2024}\natexlab{}.
\newblock \showarticletitle{Beyond Isolated Frames: Enhancing Sensor-Based Human Activity Recognition through Intra-and Inter-Frame Attention}.
\newblock \bibinfo{journal}{\emph{arXiv preprint arXiv:2405.19349}} (\bibinfo{year}{2024}).
\newblock


\bibitem[Tang et~al\mbox{.}(2021)]%
        {tang2021selfhar}
\bibfield{author}{\bibinfo{person}{Chi~Ian Tang}, \bibinfo{person}{Ignacio Perez-Pozuelo}, \bibinfo{person}{Dimitris Spathis}, \bibinfo{person}{Soren Brage}, \bibinfo{person}{Nick Wareham}, {and} \bibinfo{person}{Cecilia Mascolo}.} \bibinfo{year}{2021}\natexlab{}.
\newblock \showarticletitle{Selfhar: Improving human activity recognition through self-training with unlabeled data}.
\newblock \bibinfo{journal}{\emph{Proceedings of the ACM on interactive, mobile, wearable and ubiquitous technologies}} \bibinfo{volume}{5}, \bibinfo{number}{1} (\bibinfo{year}{2021}), \bibinfo{pages}{1--30}.
\newblock


\bibitem[Team et~al\mbox{.}(2024)]%
        {team2024gemma}
\bibfield{author}{\bibinfo{person}{Gemma Team}, \bibinfo{person}{Thomas Mesnard}, \bibinfo{person}{Cassidy Hardin}, \bibinfo{person}{Robert Dadashi}, \bibinfo{person}{Surya Bhupatiraju}, \bibinfo{person}{Shreya Pathak}, \bibinfo{person}{Laurent Sifre}, \bibinfo{person}{Morgane Rivi{\`e}re}, \bibinfo{person}{Mihir~Sanjay Kale}, \bibinfo{person}{Juliette Love}, {et~al\mbox{.}}} \bibinfo{year}{2024}\natexlab{}.
\newblock \showarticletitle{Gemma: Open models based on gemini research and technology}.
\newblock \bibinfo{journal}{\emph{arXiv preprint arXiv:2403.08295}} (\bibinfo{year}{2024}).
\newblock


\bibitem[Thukral et~al\mbox{.}(2024)]%
        {thukral2024layout}
\bibfield{author}{\bibinfo{person}{Megha Thukral}, \bibinfo{person}{Sourish~Gunesh Dhekane}, \bibinfo{person}{Shruthi~K Hiremath}, \bibinfo{person}{Harish Haresamudram}, {and} \bibinfo{person}{Thomas Ploetz}.} \bibinfo{year}{2024}\natexlab{}.
\newblock \showarticletitle{Layout Agnostic Human Activity Recognition in Smart Homes through Textual Descriptions Of Sensor Triggers (TDOST)}.
\newblock \bibinfo{journal}{\emph{arXiv preprint arXiv:2405.12368}} (\bibinfo{year}{2024}).
\newblock


\bibitem[Yu et~al\mbox{.}(2019)]%
        {yu2019novel}
\bibfield{author}{\bibinfo{person}{Yufeng Yu}, \bibinfo{person}{Yuelong Zhu}, \bibinfo{person}{Dingsheng Wan}, \bibinfo{person}{Huan Liu}, {and} \bibinfo{person}{Qun Zhao}.} \bibinfo{year}{2019}\natexlab{}.
\newblock \showarticletitle{A novel symbolic aggregate approximation for time series}. In \bibinfo{booktitle}{\emph{Proceedings of the 13th International Conference on Ubiquitous Information Management and Communication (IMCOM) 2019 13}}. Springer, \bibinfo{pages}{805--822}.
\newblock


\bibitem[Zhao et~al\mbox{.}(2023)]%
        {zhao2023survey}
\bibfield{author}{\bibinfo{person}{Wayne~Xin Zhao}, \bibinfo{person}{Kun Zhou}, \bibinfo{person}{Junyi Li}, \bibinfo{person}{Tianyi Tang}, \bibinfo{person}{Xiaolei Wang}, \bibinfo{person}{Yupeng Hou}, \bibinfo{person}{Yingqian Min}, \bibinfo{person}{Beichen Zhang}, \bibinfo{person}{Junjie Zhang}, \bibinfo{person}{Zican Dong}, {et~al\mbox{.}}} \bibinfo{year}{2023}\natexlab{}.
\newblock \showarticletitle{A survey of large language models}.
\newblock \bibinfo{journal}{\emph{arXiv preprint arXiv:2303.18223}} (\bibinfo{year}{2023}).
\newblock


\end{thebibliography}

\end{document}